\documentclass[10pt,twocolumn,letterpaper]{article}

\usepackage[pagenumbers]{cvpr} 

\usepackage[dvipsnames]{xcolor}

\definecolor{cvprblue}{rgb}{0.21,0.49,0.74}
\usepackage[pagebackref,breaklinks,colorlinks,citecolor=cvprblue]{hyperref}
\newcommand{\norm}[1]{\left\lVert#1\right\rVert}
\usepackage{xcolor,colortbl}

\definecolor{Gray}{gray}{0.3}
\definecolor{LightCyan}{rgb}{0.88,1,1}

\newcolumntype{a}{>{\columncolor{Gray}}c}
\newcolumntype{b}{>{\columncolor{white}}c}
\usepackage{booktabs}
\usepackage{graphicx}
\usepackage{array}
\newcolumntype{I}{!{\vrule width 3pt}}
\newlength\savedwidth

\newlength\savewidth
\newcommand\shline{\noalign{\global\savewidth\arrayrulewidth
\global\arrayrulewidth 1.5pt}%
\hline
\noalign{\global\arrayrulewidth\savewidth}}
\def\paperID{10167} 
\def\confName{CVPR}
\def\confYear{2024}

\title{SEEAvatar: Photorealistic Text-to-3D Avatar Generation\\ with Constrained Geometry and Appearance}

\author{Yuanyou Xu \hspace{1.5em}  Zongxin Yang \hspace{1.5em}  Yi Yang \\
ReLER, CCAI, Zhejiang University\\
{\tt\small yoxu@zju.edu.cn, zongxinyang1996@gmail.com, yangyics@zju.edu.cn}
}

\begin{document}

\pagestyle{plain}
\twocolumn[{
\renewcommand\twocolumn[1][]{#1}
\maketitle
\begin{center}
\setlength{\abovecaptionskip}{8pt} 
\setlength{\belowcaptionskip}{0pt} 
\centering

\vspace{-16pt}
\includegraphics[width=0.99\textwidth]{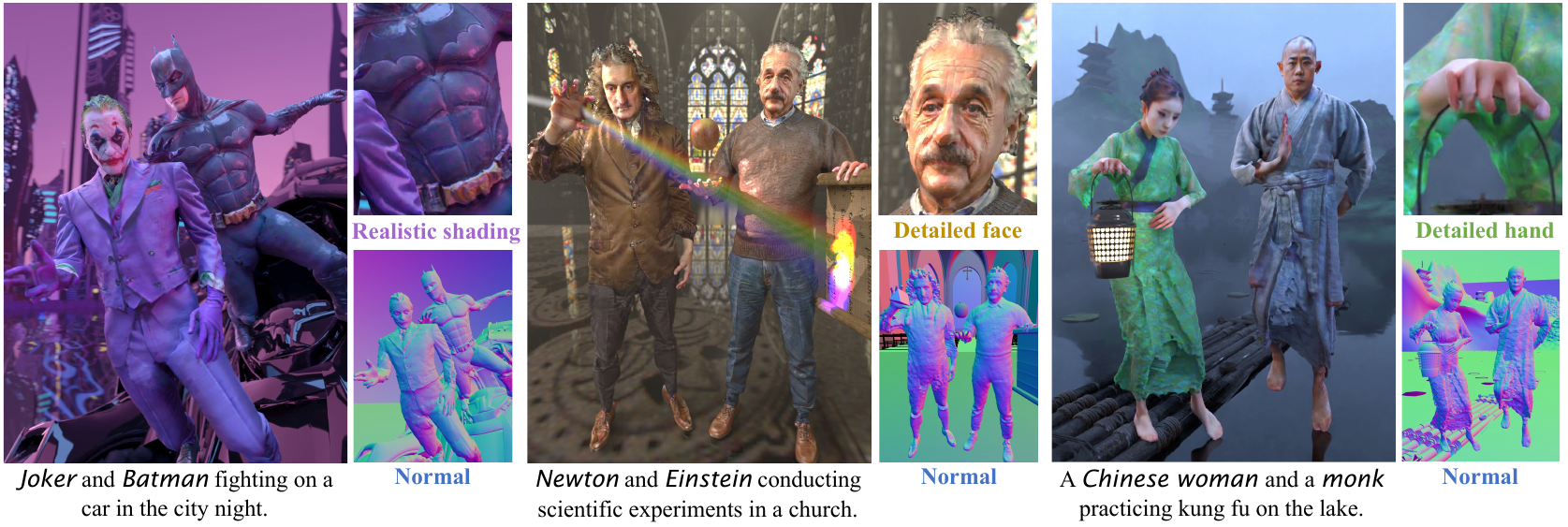}
\vspace{-8pt}
\captionof{figure}{We present three cases to demonstrate the application of the generated avatars in classic graphics workflows. In above scenes, the avatars are generated by our method and represented as meshes and textures, and exported to Blender for posing and rendering. The meshes have \textit{decent body shapes} and highly \textit{detailed structures} in hands and face. The textures have \textit{correct colors} and \textit{rich local details}. With the high quality geometry and texture assets, photorealistic portraits can be created. Additional 3D assets are used to build the scenes.}
\label{fig:open}
\end{center}}]
\begin{abstract}
Powered by large-scale text-to-image generation models, text-to-3D avatar generation has made promising progress. However, most methods fail to produce photorealistic results, limited by imprecise geometry and low-quality appearance. Towards more practical avatar generation, we present SEEAvatar, a method for generating photorealistic 3D avatars from text with SElf-Evolving constraints for decoupled geometry and appearance.
For geometry, we propose to constrain the optimized avatar in a decent global shape with a template avatar. The template avatar is initialized with human prior and can be updated by the optimized avatar periodically as an evolving template, which enables more flexible shape generation. Besides, the geometry is also constrained by the static human prior in local parts like face and hands to maintain the delicate structures.
For appearance generation, we use diffusion model enhanced by prompt engineering to guide a physically based rendering pipeline to generate realistic textures. The lightness constraint is applied on the albedo texture to suppress incorrect lighting effect. 
Experiments show that our method outperforms previous methods on both global and local geometry and appearance quality by a large margin. Since our method can produce high-quality meshes and textures, such assets can be directly applied in classic graphics pipeline for realistic rendering under any lighting condition. 
Project page at: \href{https://yoxu515.github.io/SEEAvatar/}{https://yoxu515.github.io/SEEAvatar/}.
\end{abstract}    
\section{Introduction}

\begin{figure*}[h]
\centering
\includegraphics[width=0.95\textwidth]{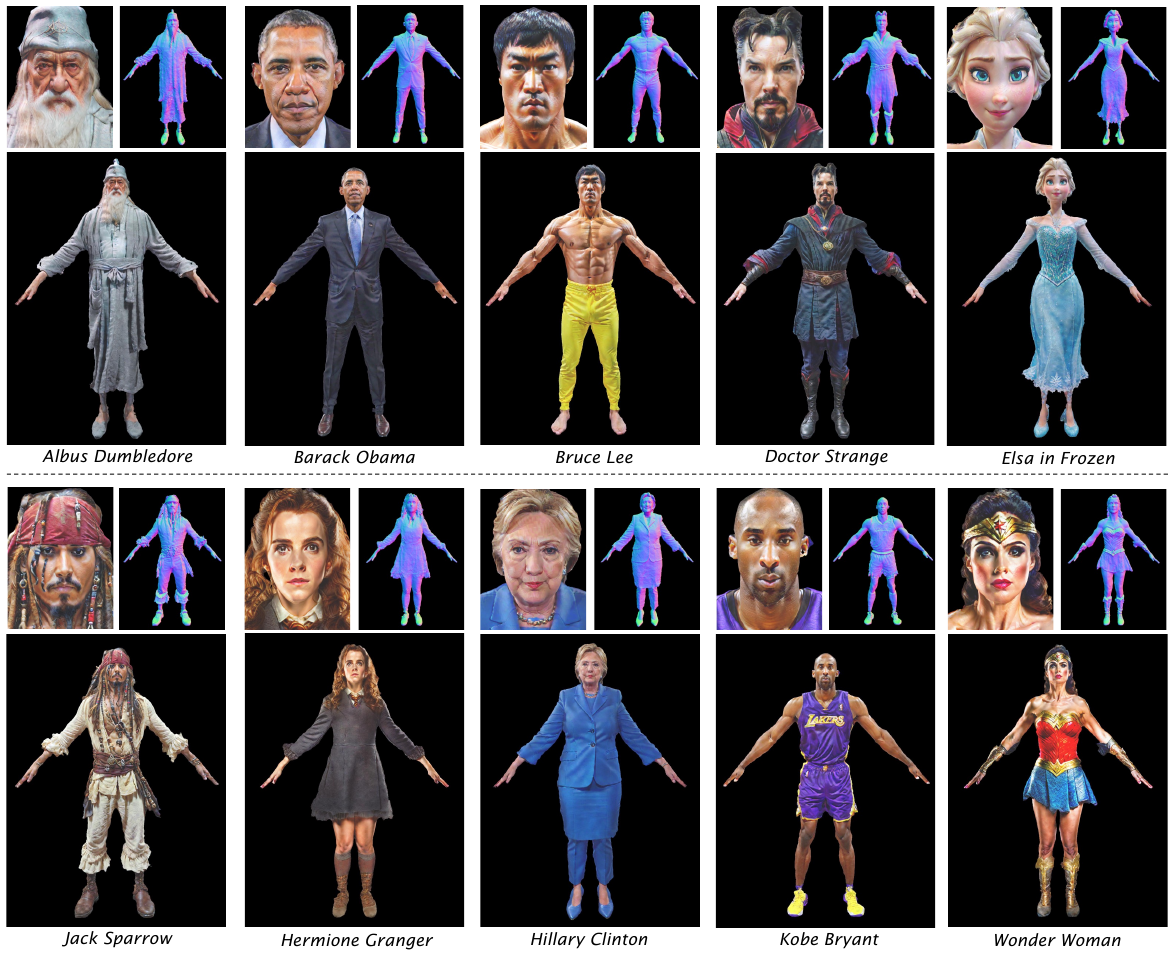}
\vspace{-8pt}
\caption{\textbf{Generated avatars from text prompts.} Full body colors and normals are rendered in the front view, and faces are rendered in a closer distance. Please \textbf{zoom in} for better view.}
\vspace{-12pt}
\label{fig:gallery}
\end{figure*}
With the development of computer vision and computer graphics techniques \cite{kajiya1986rendering,lorensen1998marching,rother2004grabcut,phong1998illumination,snavely2006photo}, creating and seeing a world from fantasy have been made possible. Although the film industry and gaming industry have developed mature workflows to create ideal visual contents and effects from 3D assets, producing photorealistic 3D assets often requires expensive scanning machines and intensive work by professional artists. Easier solutions for generating photorealistic 3D assets are meaningful in applications like virtual reality, game and film production.


Therefore, we are committed to constructing a framework for creating photorealistic 3D avatars from text. However, such a task is challenging because high realism requires both high resolution and high quality geometry and appearance. For geometry, 1) \textit{decent body proportions} and \textit{diverse clothes styles} should be correctly generated. For example, an avatar with a large head and a small body will look unnatural \cite{zeng2023avatarbooth,liao2023tada}. 2) The \textit{detailed shape structures} in hands, face and clothes should be delicately articulated. Missing hands or facial features is a severe detriment to realism \cite{cao2023dreamavatar,huang2023dreamwaltz}. For appearance, 1) \textit{correct color style} and \textit{high resolution details} are necessary for visually pleasing results. Over-saturated colors \cite{huang2023avatarfusion,cao2023dreamavatar} or low resolution details \cite{kolotouros2023dreamhuman} will result in unrealistic results. 2) In order to achieve photorealistic results under any light conditions, explicit meshes and textures are needed for \textit{physically based rendering} (PBR) in classic graphics pipeline. Avatars represented in neural fields with entangled color and geometry \cite{mildenhall2021nerf} are hard to be applied into classic tools and workflows for further application like relighting \cite{cao2023dreamavatar,huang2023dreamwaltz,kolotouros2023dreamhuman}.
A recent work Fantasia3D \cite{chen2023fantasia3d} for 3D content generation decouples geometry and appearance. However, it fails to generate high quality 3D avatars from text. The normal-based guidance for geometry modeling lacks stability and fails to generate avatars with fine human shapes and well-structured details. In addition, the PBR pipeline guided by the diffusion model in appearance modeling often absorbs lighting into albedo textures. These problems hinder Fantasia3D from producing favorable results for the avatar generation task.


In order to tackle above challenges, we present SEEAvatar for photorealistic avatar generation from text. We follow the decoupled framework from \cite{chen2023fantasia3d}, and further improve the geometry and appearance generation with SElf-Evolving constraints. For geometry generation, we set up constraints for both global shape and local structures. More specifically, we set two model avatars, one is the \textit{current avatar} and the other is the \textit{template avatar}. The current avatar is represented as DMTet \cite{shen2021deep}, which is guided by 2D diffusion model. The template avatar is responsible for constraining the current avatar in a decent human shape. We apply signed distance function (SDF) constraint for shape controlling and normal constraint for surface smoothing between the template avatar and current avatar. The simplest way is to set the template avatar constantly as the human prior SMPL-X \cite{loper2023smpl,pavlakos2019expressive}. However, this will impede generating shapes which are different from the original prior, for example, a woman in a dress (\cref{fig:ablation_geo} (c)). To make the generation more flexible, we loosen the constraint by updating the template avatar periodically. This can enhance the potential to generate more diverse shapes. Although the generation becomes flexible, delicate local structures like hands will also be easier to be ruined. Therefore, we further employ the static human prior for local constraints. The combination of the evolving template avatar and the static human prior can produce avatars with decent body proportion, flexible clothes style and delicate local structures.

For appearance generation, we first employ prompt engineering to enhance the diffusion model, which has been widely used by the open community \cite{Civitai,HuggingFace} but was often ignored by prior works. Proper positive prompt and negative prompt can largely improve the appearance quality and enrich the details. Although the quality is improved, we find the diffusion model tends to generate fancy lighting effects and it would be hard for the model to decompose the lighting from the albedo. A key observation is that in early steps, the generated albedo is close to pure colors (\cref{fig:lightness}). Similar to the geometry constraints, we also set an evolving template avatar for appearance generation. We constrain the lightness of the current avatar's albedo by the template albedo, which can suppress the incorrect light effects.

In summary, our method is able to generate photorealistic avatars under geometry and appearance constraints. Our contributions are as following:
\begin{itemize}
    \item We incorporate the decoupled geometry and appearance framework for photorealistic avatar generation. The generated avatars are represented by high quality meshes and textures, which are friendly to classic graphics pipelines for further applications.
    \item For the geometry generation, we propose to constrain the current avatar by an evolving template avatar. Both global and local constraints are applied on SDFs and normals, which enable us to generate avatars with decent global shapes and fine local structures.
    \item For appearance generation, we enhance the diffusion model with prompt engineering. Besides, we propose to constrain the lightness of current avatar's albedo by the evolving template, which produces high quality PBR textures with less lighting involved.
\end{itemize}

\section{Related work}
\label{sec:related}
\textbf{Text-to-3D generation}
A common way for 3D content generation methods is to guide 3D representations by 2D models trained on image and text pairs. Early methods like CLIP-Mesh \cite{mohammad2022clip}, CLIP-forge \cite{sanghi2022clip} and DreamFields \cite{jain2022zero} use CLIP \cite{radford2021learning} as guidance. Later, diffusion models \cite{ho2020denoising,song2020denoising,ramesh2021zero,rombach2022high,saharia2022photorealistic} have shown great potential in image generation. DreamFusion \cite{poole2022dreamfusion} propose to distill 2D diffusion models into 3D neural fields with score distillation sampling (SDS). Score Jacobian Chaining (SJC) \cite{wang2023score} follows the similar idea with a different formulation. Magic3D \cite{lin2023magic3d} proposes a two stage method combing Nerf with mesh fine-tuning. ProlificDreamer \cite{wang2023prolificdreamer} improves the generation quality by Variational Score Distillation (VSD). Fantasia3D \cite{chen2023fantasia3d} decouples geometry and appearance and is able to generate explicit meshes and textures. Generating textures on given meshes has also been studied by some works \cite{chen2023text2tex,richardson2023texture,siddiqui2022texturify}. Although these methods can generate diverse objects, they are not delicately designed for avatar generation. We incorporate the decoupled framework from \cite{chen2023fantasia3d}, and specialize in photorealistic geometry and appearance generation.


\noindent \textbf{Text-to-3D avatar generation} AvatarCLIP \cite{hong2022avatarclip} uses CLIP loss to guide distance and color fields \cite{wang2021neus} for avatar generation. DreamAvatar \cite{cao2023dreamavatar} uses diffusion model as guidance and optimize dual space neural fields to realize pose control. AvatarCraft \cite{jiang2023avatarcraft} propose implicit neural representation with controllable shapes and poses. DreamWaltz \cite{huang2023dreamwaltz} incorporates the pose conditioned ControlNet \cite{zhang2023adding} to optimize an animatable avatar representation. DreamHuman \cite{kolotouros2023dreamhuman} uses imGHUM \cite{alldieck2021imghum} as human prior for pose control and generates human avatars with semantic zoom of multiple body parts. AvatarFusion \cite{huang2023avatarfusion} generates avatars with separate body and clothes neural fields and renders by fusing the representations. AvatarVerse \cite{zhang2023avatarverse} trains a dense pose based ControlNet and generates avatars with progressive training. TADA \cite{liao2023tada} directly optimizes the displacement and color for the human prior meshes, and achieves fully animatable avatars. Some methods also focus on head \cite{zhang2023dreamface,han2023headsculpt} or upper body \cite{zhang2023text} generation. Most of these methods use entangled color and geometry neural fields, which makes them hard to be applied in classic graphics pipelines for realistic rendering under any lighting condition. In addition, volumtric rendering can be heavy in memory consumption, which prevents them from rendering high resolution images.

\section{Method}
\begin{figure*}[h]
\centering
\includegraphics[width=0.9\textwidth]{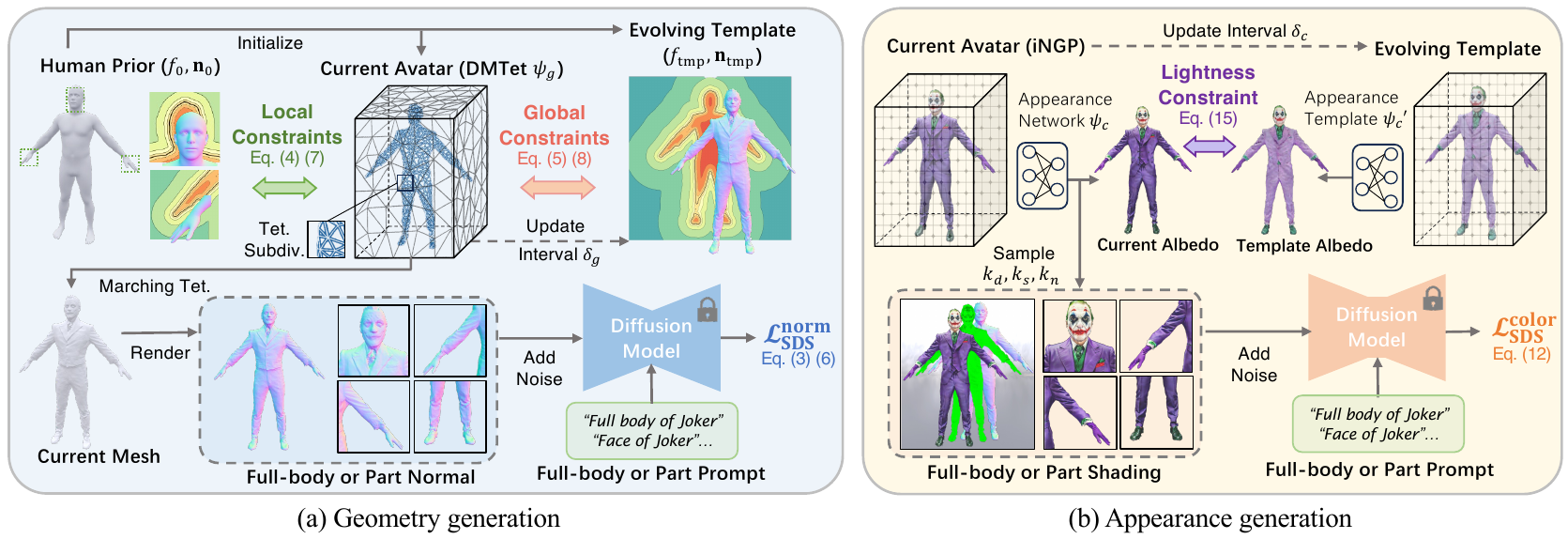}
\vspace{-8pt}
\caption{\textbf{Overview}, consisting of geometry (\S\ref{sec:geo}) and appearance (\S\ref{sec:apr}) generation. For geometry generation, we use DMTet \cite{shen2021deep} as the 3D shape representation, and optimize it by normal-based SDS loss. The global SDF and normal constraints are applied between current avatar and the evolving template, and local constraints are from the static human prior. For appearance, we use iNGP \cite{muller2022instant} to represent the PBR texture field. Albedo/roughness/normal are sampled from the appearance representation for shading and optimized by color SDS loss. The lightness constraint is applied between the current albedo and the template albedo to suppress lighting effects.}
\vspace{-12pt}
\label{fig:method}
\end{figure*}
\subsection{Preliminary}
\textbf{Score distillation sampling (SDS)} SDS is proposed in DreamFusion as a loss for optimizing 3D representation by a 2D diffusion model. A 2D image can be rendered from 3D representation as $x=g(\theta)$. The SDS loss minimizes the difference between the predicted noise $\epsilon_\phi$ with the added random noise $\epsilon$:
\begin{equation}
    \nabla_{\theta} \mathcal{L}_{\mathrm{SDS}}(\theta,\mathbf{x})= \mathbb{E}_{t,\epsilon}[w(t)(\epsilon_{\phi}(\mathbf{z_t};y,t)-\epsilon)\frac{\partial \mathbf{x}}{\partial \theta}]
\end{equation}
where $\mathbf{z_t}$ is the noised image, $t$ is the time step, $y$ is the text condition, and $w(t)$ is a weighting function determined by the time step.

\noindent\textbf{SMPL-X} 
SMPL is proposed in \cite{loper2023smpl} as a parametric human model. SMPL-X \cite{pavlakos2019expressive} is an extension of SMPL by integrating head \cite{li2017learning} and hand \cite{romero2022embodied} models. SMPL-X uses standard vertex-based linear blend skinning with blend shapes, parameterized by shape, pose and facial expression parameters. We use the semantic label of the vertices to 1) extract body parts like face or hands for our local constraints, 2) localize camera poses for local part render.

\noindent\textbf{DMTet} Deep Marching Tetrahedra \cite{gao2020learning,shen2021deep} is proposed for high resolution 3D shape synthesis, as a hybrid representation of a tetrahedral grid and an implicit sign distance function (SDF). The SDF can be efficiently represented by multi-resolution hash encoding \cite{muller2022instant} with a MLP network. Explicit mesh can be extracted from the tetrahedral grid by Marching Tetrahedra (MT) in a differentiable manner.

\subsection{Geometry generation}
\label{sec:geo}
\subsubsection{Human prior optimization}
Given a text description $y$ of the target avatar, we start from optimizing the shape of the human prior. The mesh can be derived from SMPL-X with shape parameters $\beta$. The normal image is obtained by differentiable rendering. We guide the optimization of $\beta$ by the diffusion model by SDS loss. In this stage, the basic body shape of the avatar is determined, for example, fat or thin. After optimizing shape parameters, the human prior is used for the following stages.

\vspace{-8pt}
\subsubsection{Initialization}
We set two avatars during optimization: one is the current avatar, the other is the template avatar. The current avatar is represented by DMTet and its SDF $f_{\mathrm{cur}}$ is parameterized as a MLP network $\psi_g$, while the template avatar and its SDF $f_{\mathrm{tmp}}$ are initialized by the human prior mesh and its SDF $f_0$. Note there is no parameter for the template avatar.

At the beginning, we attempt to align the current avatar with the human prior. We sample points $\mathrm{P}=\{p_i\in \mathbb{R}^3\}$ around the mesh surface together with some random points and optimize the following loss:
\begin{equation}
\label{eq:init}
    \mathcal{L}^{\mathrm{init}}_{\mathrm{SDF}} = \norm{f_{\mathrm{cur}}(\mathrm{P})-f_{0}(\mathrm{P})}^2_2
    =\sum_{p_i\in \mathrm{P}}{\norm{f_{\mathrm{cur}}(p_i)-f_{0}(p_i)}^2_2}.
\end{equation}
After initialization, the current avatar and template avatar are both aligned with the human prior.

\vspace{-5pt}
\subsubsection{Geometry deformation stage}
In the coarse stage, the mesh of the current avatar is extracted from DMTet by Marching Tetrahedra. Then the mesh is passed to the differentiable renderer. The normal image $\mathbf{n}\in \mathbb{R}^{h\times w\times3}$ and the mask $\mathbf{a}\in \mathbb{R}^{h\times w\times1}$ are rendered by a sampled camera pose (details are in \cref{sec:implement}). Then they are concatenated and scaled into a smaller size by interpolation as $\mathbf{n_a}\in \mathbb{R}^{h'\times w'\times4}$. Noises are directly added to the $\mathbf{n_a}$ for SDS loss:
\begin{equation}
    \nabla_{\psi_g} \mathcal{L}_{\mathrm{SDS}}^{\mathrm{norm}}(\phi,\mathbf{n_a})= \mathbb{E}_{t,\epsilon}[w(t)(\epsilon_{\phi}(\mathbf{z_t^{n_a}};y,t)-\epsilon)\frac{\partial \mathbf{n_a}}{\partial \psi_g}]
\end{equation}
In this stage, the avatar's global shape will be determined in a coarse manner. Since the guidance is directly on the normal space but not latent space, this geometry deformation could be intense but unstable and easy to deviate from the correct shape and lose local structures from human prior.

\noindent\textbf{Global evolving SDF constraint}
In order to keep the geometry in a decent human shape during optimization, we impose SDF constraint on the current avatar from the template avatar. Specifically, we constantly optimize the global template loss during generation:
\begin{equation}
\label{eq:tmp}
    \mathcal{L}^{\mathrm{glb}}_{\mathrm{SDF}} = \norm{f_{\mathrm{cur}}(\mathrm{P})-f_{\mathrm{tmp}}(\mathrm{P})}^2_2.
\end{equation}
The full formulation is similar to \cref{eq:init}.
Although the constraint can avoid severe deformation from the human prior, it will also sacrifice the generation flexibility and it will become hard to generate diverse geometry shapes. To alleviate the issue, we update the template avatar periodically as the self-evolving constraint $f_{\mathrm{cur}}\Rightarrow f_{\mathrm{ref}}$. More specifically, the mesh of the current avatar is extracted in every $\delta_g$ step, and the mesh is converted to SDF as the new template.

\noindent\textbf{Local static SDF constraint}
An evolving template avatar helps to loosen the constraint, but some delicate local structures from human prior may also be ruined due to the instability of the coarse stage SDS optimization. In order to keep the detailed structures of the face and hands, we further employ a local template loss:
\begin{equation}
\label{eq:sdf_loc}
    \mathcal{L}_{\mathrm{SDF}}^{\mathrm{loc}} = \sum_i{w_i\norm{f_{\mathrm{cur}}(\mathrm{Q}_i)-f_{0}(\mathrm{Q}_i)}^2_2}
\end{equation}
where $f_0$ is the SDF of the original human prior mesh, and $\mathrm{Q}_i$ is sampled point set for the $i$-th local part. Since SMPL-X has semantic labels, we select face, hands and feet as local template parts. Feet are mainly for avatars with bare feet. The loss of all parts are summed with weights $w_i$.

\subsubsection{Geometry refining stage}
In refining stage, the normal image $\mathbf{n}$ is mapped to latent space by the auto-encoder as $\mathbf{z^n}$ before adding noise for SDS loss:
\begin{equation}
\small
    \nabla_{\psi_g} \mathcal{L}_{\mathrm{SDS}}^{\mathrm{norm}}(\phi,\mathbf{n})= \mathbb{E}_{t,\epsilon}[w(t)(\epsilon_{\phi}(\mathbf{z_t^n};y,t)-\epsilon)\frac{\partial \mathbf{z^n}}{\partial \mathbf{n}}\frac{\partial \mathbf{n}}{\partial \psi_g}].
\end{equation}
Different from the coarse stage, the refining stage will not cause large deformation but generate detailed shapes like wrinkles in clothes. However, the initial resolution of DMTet in coarse stage is set as 256, which might be insufficient to express fine details in high quality.

\noindent\textbf{Tet subdivision}  In order to increase the resolution, we subdivision the tetrahedral grid during refineing stage. In previous works \cite{shen2021deep,munkberg2022extracting,huang2023tech}, the subdivision is performed only in tetrahedra intersecting with the surface. However, we find such a strategy may cause discontinuous faces in the mesh. So we subdivide the tetrahedral grid around the surface in a fixed interval. In detail, we select the tetrahedra with mean vertex SDF value smaller than 0.2. The local subdivision will only increase a little GPU memory consumption, but the tetrahedra become $8\times$ around the surface.

\noindent\textbf{Global and local normal constraints} As the resolution becomes high, the geometry will exhibit noisy and bumpy characteristics due to the large number of vertices. Inspired by \cite{huang2023tech}, we employ normals to smooth the surface. Similar to our SDF constraints, normal constraints are imposed both globally and locally on the current avatar by the template and prior:
\begin{equation}
    \mathcal{L}_{\mathrm{norm}}^{\mathrm{glb}} = \norm{\mathbf{n}_{\mathrm{cur}}-\mathbf{n}_{\mathrm{tmp}}}^2_2,
\end{equation}
\begin{equation}
    \mathcal{L}_{\mathrm{norm}}^{\mathrm{loc}} = \sum_i{k_i\norm{\mathbf{n}_{\mathrm{cur}}-\mathbf{n}_{0}^i}^2_2}.
\end{equation}
where $\mathbf{n}_{\mathrm{cur}}$/$\mathbf{n}_{\mathrm{tmp}}$ is the normal image rendered from the current/template avatar, and $\mathbf{n}_{0}^i$ is rendered by only a part mesh from the human prior. $k_i$ is the weight for the $i$-th part. We mainly use face for this normal constraint since it is the most important local part. In summary, during the geometry generation, the total loss consists of SDS loss, SDF loss and normal loss. Both the SDF loss and normal loss include global and local terms:
\begin{align}
    \mathcal{L}_{\mathrm{geo}} &= \lambda_{\mathrm{SDS}}\mathcal{L}_{\mathrm{SDS}}^{\mathrm{norm}} + 
    \mathcal{L}_{\mathrm{SDF}} + \mathcal{L}_{\mathrm{norm}}\\
    \mathcal{L}_{\mathrm{SDF}} &= \alpha_r\mathcal{L}_{\mathrm{SDF}}^{\mathrm{glb}} + \alpha_l\mathcal{L}_{\mathrm{SDF}}^{\mathrm{loc}}\\
    \mathcal{L}_{\mathrm{\mathrm{norm}}} &= \beta_r\mathcal{L}_{\mathrm{norm}}^{\mathrm{glb}} + \beta_l\mathcal{L}_{\mathrm{norm}}^{\mathrm{loc}}
\end{align}

\subsection{Appearance generation}
\label{sec:apr}
Given the mesh generated in geometry stages, we then generate appearance for it. The appearance is represented as a neural field with multi-resolution hash encoding \cite{muller2022instant} parameterized by a MLP $\psi_c$. For the query points from the generated mesh, the MLP network predicts diffuse term $k_d$ as albedo color, specular term $k_s$ including roughness and metalness, and the normal term $k_n$ following PBR workflow. With these textures, the RGB image $\mathbf{x}$ is rendered with a differentiable renderer \cite{munkberg2022extracting,hasselgren2022shape}, given a HDRI as the environment light. Then, the RGB image is guided by the diffusion model by SDS loss:
\begin{equation}
\small
    \nabla_{\psi_c} \mathcal{L}_{\mathrm{SDS}}^{\mathrm{color}}(\phi,\mathbf{x})= \mathbb{E}_{t,\epsilon}[w(t)(\epsilon_{\phi}(\mathbf{z_t^x};y,t)-\epsilon)\frac{\partial \mathbf{z^x}}{\partial \mathbf{x}}\frac{\partial \mathbf{x}}{\partial \psi_c}].
\end{equation}

\subsubsection{Uniform scaling}
 The standard normalization for the query points is non-uniform or anisotropic, i.e. points normalized to $[0,1]^3$ by $(p-p_{\mathrm{min}})/(p_{\mathrm{max}}-p_{\mathrm{min}})$, given the corners of the 3D bounding box of all points as $p_{\mathrm{max}},p_{\mathrm{min}}\in \mathbb{R}^3$. For a point $p(x,y,z)$, the scaled factors for each dimension will be $(s_x,s_y,s_z)=p_{\mathrm{max}}-p_{\mathrm{min}}$. We find this operation will result in artifacts in the texture along the depth dimension (\cref{fig:ablation_ap} (a)). The scale of the avatar along the depth dimension is relatively small compared with wideness and height. Therefore, we use a uniform scaling which produces no artifacts along the depth dimension:
\begin{equation}
    p'= (p-p_{\mathrm{min}})/s,\quad
    s = \max(s_x,s_y,s_z).
\end{equation}


\subsubsection{Self-evolving lightness constraint}
\noindent \textbf{Model enhancement} In order to generate appearance with richer details, we use prompt engineering to enhance the diffusion model. The technique is widely used in practice by the community, but we notice prior works often use simple prompt, which may not fully activate the potential of the model. Proper positive prompt and negative prompt can largely improve the appearance quality. More details can be found in \cref{sec:implement}. 

\begin{figure}[t]
\centering
\includegraphics[width=0.95\columnwidth]{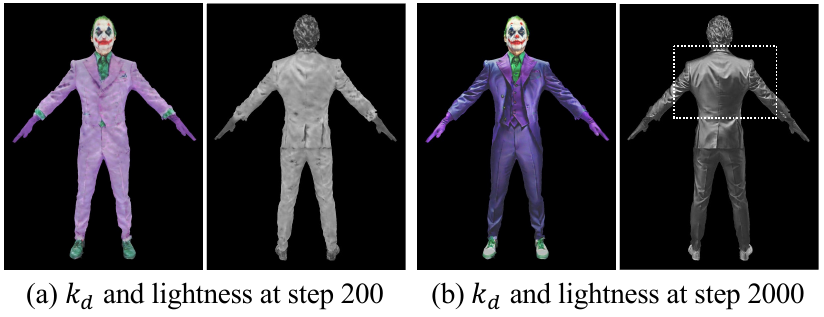}
\vspace{-8pt}
\caption{\textbf{Albedo and lightness} at early step and late step (\S\ref{sec:light}).}
\vspace{-12pt}
\label{fig:lightness}
\end{figure}

\label{sec:light}
\noindent \textbf{Lighting effects} Although the quality is improved with the help of prompt engineering, we also observe that the diffusion model tends to produce fancy lighting effects which is inconsistent with the environment lighting. As a result, the differentiable rendering system may absorb the lighting effects into albedo colors. 
An important empirical observation is that the albedo colors do not include many lighting effects in early steps, as shown in \cref{fig:lightness}. We find the generation process follows the coarse to fine rule: pure colors are first generated for each part, and local details and lighting effects are generated in later steps. Therefore, our idea is to constrain late steps by early steps.

\noindent \textbf{Lightness constraint} Similar to the geometry generation, we set up the current avatar with a template avatar. The template avatar's appearance is initialized by the current avatar in an early step. We constrain the distance between the lightness (or luminance) of the current avatar and template avatar. However, directly constraining the current avatar with the template avatar may prevent current avatar from generating fine details. To alleviate the problem, the template avatar's appearance network $\psi_c'$ is updated by the current avatar's network $\psi_c$ every $\delta_c$ steps. In addition, we scale the albedo images into a smaller scale, which will make the constraint only effective on coarse lighting cues and loose on local details. Specifically, we optimize the following loss for lightness constraint:
\begin{align}
    Y(I) &= (I_r+I_g+I_b)/3,\\
    \mathcal{L}_{\mathrm{lgt}} &= \norm{S(Y(k_d^{\mathrm{cur}}))-S(Y(k_d^{\mathrm{tmp}}))}_2^2
\end{align}
where $k_d^{\mathrm{cur}}$/$k_d^{\mathrm{tmp}}$ is the rendered albedo of current/template avatar, and $S$ is the down-scaling function.
\section{Experiment}
\begin{figure*}[t]
\centering
\includegraphics[width=0.95\textwidth]{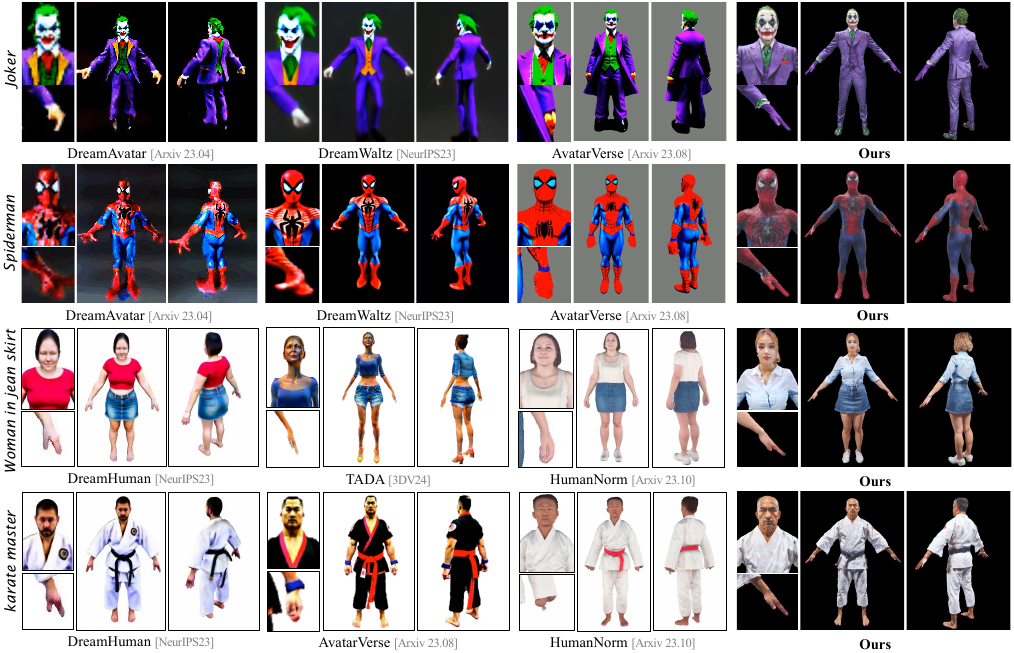}
\vspace{-8pt}
\caption{\textbf{Qualitative comparison} (\S\ref{sec:qua}). We compare our method with DreamAvatar \cite{cao2023dreamavatar}, DreamWaltz \cite{huang2023dreamwaltz}, AvatarVerse \cite{zhang2023avatarverse}, DreamHuman \cite{kolotouros2023dreamhuman}, TADA \cite{liao2023tada} and HumanNorm \cite{huang2023humannorm}. Front and back views are presented, and faces and hands are cropped and enlarged for detailed comparison. Please zoom in for better view.}
\vspace{-12pt}
\label{fig:compare}
\end{figure*}

\subsection{Implementation details}
\label{sec:implement}
\textbf{Camera sampling}
Similar to DreamHuman \cite{kolotouros2023dreamhuman}, we sample camera poses for multiple parts of the body. We obtain semantic labels from SMPL-X \cite{pavlakos2019expressive} and set head, arms and feet as separate parts. In each training step, the full body or one part is randomly chosen as the center of the view and the camera is also randomly rotated for rendering. The prompt patterns are like \textit{``A full length DSLR photo of ..."} for full body and \textit{``A DSLR photo of ...'s ..."} for local parts. 


\noindent\textbf{Diffusion models} For geometry generation, we use Stable Diffusion 2.1 base (SD2.1b) \cite{StableDiffusionv2.1-base}.
For appearance generation, we use Realistic Vision 5.1 (RV5.1) \cite{RealisticVision} by default. SD2.1b is also tested but we find RV5.1 can generate more visually pleasing appearance. More details and examples can be found in appendices.

\noindent\textbf{Prompt engineering} Using proper prompt is crucial for generating high quality 2D images from diffusion models. We take in the experience from the community \cite{Civitai,HuggingFace} and add auxiliary positive prompt and negative prompt during appearance generation.
For \textit{positive prompt}, we use ``masterpiece, Studio Quality, 8k, ultra-HD, next generation" to enhance the generation quality. The positive prompt is appended behind the prompt for target avatars.
For \textit{negative prompt}, the following categories are concerned: color (``noise,pattern, strange color..."), structure (``poorly drawn face, mutation,ugly...") and quality (``low quality, lowres, error..."). For historical avatars like Lincoln, adding ``sculpture,statue" will prevent the model from generating grayscale appearance.

\noindent\textbf{Generation} For geometry generation, the normal images are rendered in $512\times 512$. The update interval of template avatar is set as 5000 by default. A smaller interval can be set for complex avatars like long hair or long dress. For appearance generation, the RGB images are rendered in $768\times 768$, and final textures are sampled from the 3D representation and saved in $4096\times 4096$. The initial step of the template avatar is 400 and update interval is set as 200.
We use four Nvidia RTX 3090 GPUs with 24GB memory for all experiments. The generation process includes 9k steps for geometry and 2k steps for appearance, which takes around 2 hours. More details can be found in appendices.

\subsection{Qualitative comparison}
\label{sec:qua}
The qualitative comparison results are shown in \cref{fig:compare}. For well-known iconic characters, we compare our method with DreamAvatar \cite{cao2023dreamavatar}, DreamWaltz \cite{huang2023dreamwaltz} and AvatarVerse \cite{zhang2023avatarverse}. The target avatars are ``\textit{SpiderMan}" and ``\textit{Joker}". For custom characters, we compare our method with DreamHuman \cite{kolotouros2023dreamhuman}, AvatarVerse \cite{zhang2023avatarverse}, TADA \cite{liao2023tada} and HumanNorm \cite{huang2023humannorm}. The target avatars are ``\textit{A karate master}" and ``\textit{A woman in jean skirt}".

From the perspective of geometry quality, our results are better on both the global shape and local structures. On the one hand, avatars are more strictly aligned with the human prior due to global geometry constraints, so the body ratios are more natural. Local structures in face and hands/feet are also well preserved thanks to the local geometry constraints. On the other hand, with the help of the evolving template, shapes like loose karate uniform or skirt can also be generated with fine details. 

As for appearance, our avatars are equipped with high resolution textures with rich details including the facial features and clothing, without any over-saturation. Photorealistic images can be produced with classic graphics pipelines with our high quality textures.

\begin{table}[t!]
\centering
\small
\setlength{\tabcolsep}{4.2pt}
\renewcommand{\arraystretch}{0.8}
\begin{tabular}{@{}rl|cccc|c@{}}
\shline
\rowcolor[HTML]{E0E0E0} 
\multicolumn{2}{c|}{Method}             & \multicolumn{1}{l}{Geo.$^G$} & \multicolumn{1}{l}{Geo.$^L$} & \multicolumn{1}{l}{Appr.$^G$} & \multicolumn{1}{l|}{Appr.$^L$} & \multicolumn{1}{c}{Avg.} \\ \hline\hline
\hspace{1mm}DreamWaltz & \cite{huang2023dreamwaltz} & 2.10                         & 1.72                         & 1.92                          & 1.72                           & 1.86        \hspace{1mm}             \\
TADA       & \cite{liao2023tada}        & 2.88                         & 2.71                         & 2.77                          & 2.69                           & 2.76             \hspace{1mm}        \\ \hline
\textbf{Ours} &     & \textbf{4.39}                & \textbf{4.36}                & \textbf{4.31}                 & \textbf{4.38}                  & \textbf{4.36}   \hspace{1mm}         \\ \shline
\end{tabular}%
\vspace{-4pt}
\caption{\textbf{Quantitative comparison} (\S\ref{sec:qua}). We compare our method with recent works by human evaluation. Both geometry (Geo.) and appearance (Appr.) are evaluated in global ($G$) and local ($L$) aspects. Higher score is better, ranging from 1 to 5.}
\vspace{-12pt}
\label{tab:user}
\end{table}

\begin{figure}[]
\centering
\includegraphics[width=0.95\columnwidth]{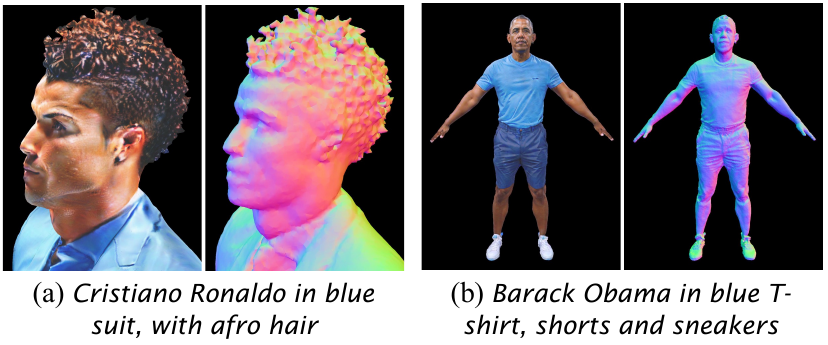}
\vspace{-8pt}
\caption{\textbf{Geometry and appearance editing} by text prompts (\S\ref{sec:edit}), such as different hair, clothes styles, colors, \textit{etc}.}
\vspace{-12pt}
\label{fig:edit}
\end{figure}

\subsection{Quantitative evaluation}
We sampled 25 avatar prompts and invited 15 volunteers for human evaluation. We compare our method with nerf-based DreamWaltz \cite{huang2023dreamwaltz} and mesh-based TADA \cite{liao2023tada}, which have available source code. The generated avatars are rendered in videos and volunteers are asked to give scores from 1 to 5 in four aspects: global shapes (body proportion, clothes style), local structures (face, hands), global appearance (color style) and local appearance (texture details). The results are in \cref{tab:user}. Our method has the highest scores in all aspects, especially in local geometry and appearance.

\subsection{Applications}
\label{sec:edit}
Since the geometry and appearance is decoupled in our framework, it would be more flexible for geometry and appearance editing. Examples are shown in \cref{fig:edit}. Avatar's body shape, clothing, hair style with different colors can be edited with different text prompts.
In addition, our method generates avatars with meshes and textures, which can be easily exported to graphics tools like Blender for further application like relighting and posing. Demos are in \cref{fig:open}.
\section{Ablation Study}
\subsection{Geometry generation}
\label{sec:geo_ablation}
\begin{figure}[t]
\centering
\includegraphics[width=0.95\columnwidth]{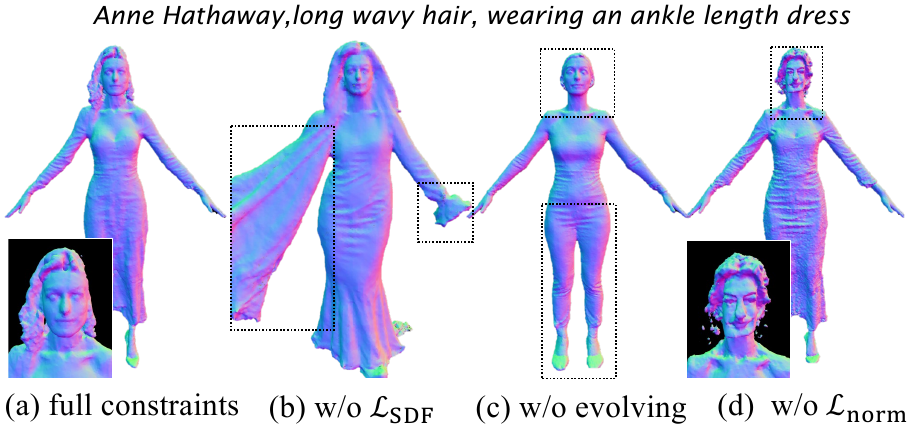}
\vspace{-8pt}
\caption{\textbf{Ablation study of geometry generation} (\S\ref{sec:geo_ablation}). The effect of SDF/normal constraint and evolving template are shown.}
\vspace{-12pt}
\label{fig:ablation_geo}
\end{figure}
The ablation results of the geometric constraints are shown in \cref{fig:ablation_geo}. The prompt is ``Anne Hathaway, long wavy hair, wearing an ankle length dress". In (a) and (b), it is demonstrated that the SDF constraint from the template avatar is crucial for generating global and local shapes conforming to the human prior. If SDF constraints are missing, generating with only normal constraints will not produce good geometry because the depth is not controlled. In addition, an evolving template is also important, shown in (a) and (c). If the template avatar is not evolving, the generated geometry can only be close to the human prior, noticing the dress and hair are not generated correctly as the prompt requires. Tet subdivision pushes the representation to a higher resolution, but it will also cause noisy surfaces like (d), if the normal constraints are not used.  

\subsection{Appearance generation}
\label{sec:apr_ablation}
\begin{figure}[t]
\centering
\includegraphics[width=0.95\columnwidth]{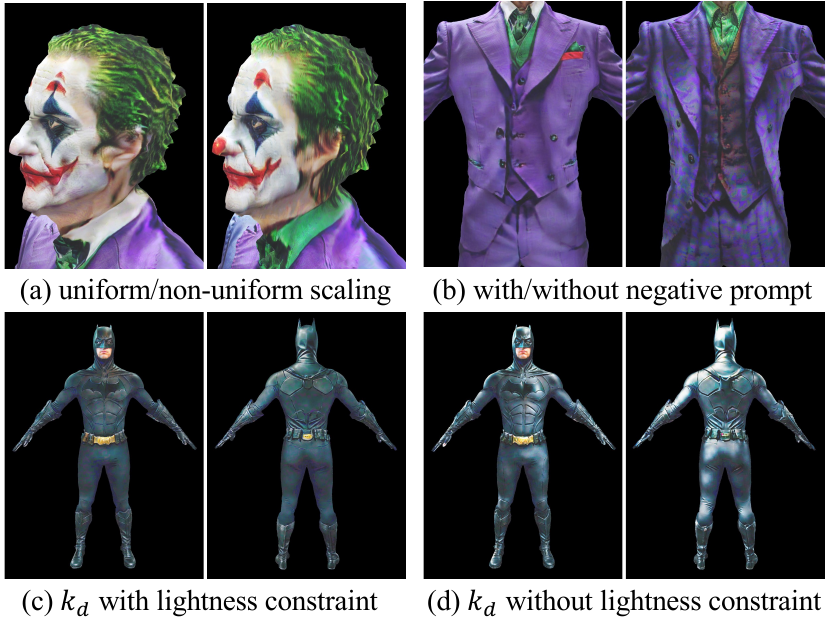}
\vspace{-8pt}
\caption{\textbf{Ablation study of appearance generation} (\S\ref{sec:apr_ablation}). The effect of uniform scaling, negative prompt and the lightness constraint are shown. Please \textbf{zoom in} for better view.}
\vspace{-12pt}
\label{fig:ablation_ap}
\end{figure}
The ablation results of the appearance generation are shown in \cref{fig:ablation_ap}. (a) shows that non-uniform scaling will result in striped artifacts in depth dimension, and uniform scaling can solve the issue. (b) demonstrates that the negative prompt can reduce over-saturation and noisy artifacts. In (c) and (d), if the constraint is removed, the guidance diffusion model tends to generate strong light effects, and fake speculars will be ``baked" into the albedo ($k_d$). With the help of our constraint, the specular parts in the albedo are suppressed and the overall lightness is more uniform.



\section{Limitations and future work}
Although high quality avatars can be generated by our method, there are still some limitations. Examples can be found in appendices. For geometry, some highly detailed structures are hard to be represented, such as hair strands and eyelashes. In addition, very loose clothes or complex accessories can not be well generated in our framework. More delicate representations may be designed to solve these issues.

For appearance, despite we follow the PBR workflow, the roughness values may not be accurately generated. In \cref{fig:open}, sharp specular can be observed on Einstein's sweater, which are caused by a too low roughness value. Besides, there are still some lighting and shadows baked into albedo colors. Our constraint is an empirical strategy by leveraging the property of the model itself. Introducing extra data as prior knowledge may help with the problem \cite{xu2023matlaber}. In addition, the generated appearance may not be exactly aligned with the geometry. This problem may be alleviated by adding more controls in appearance guidance \cite{liao2023tada,huang2023humannorm}.
\section{Conclusion}
In this work, we present SEEAvatar, a method for photorealistic avatar generation with constrainted geometry and appearance. The proposed constraints for geometry are able to control the avatar in a decent global human shape with flexible clothes styles, while maintain detailed local structures from the human prior. The SDF and normal constraints also help the geometry to deform with smooth surfaces in the subdivided high resolution representation. The proposed lightness constraint for appearance generation can effectively suppress the lighting effect in albedo colors. As a result, the high quality 3D meshes and textures generated by our method can be applied in classic workflows for photorealistic rendering.
{
    \small
    \bibliographystyle{ieeenat_fullname}
    \bibliography{main}
}

\clearpage
\setcounter{page}{1}

\maketitlesupplementary


\section{Additional qualitative comparison results}
\begin{figure*}[h]
\centering
\includegraphics[width=0.95\textwidth]{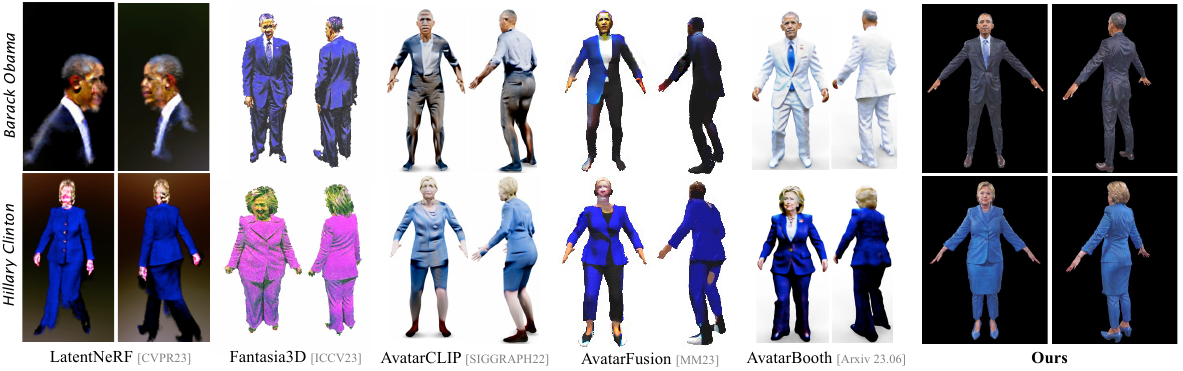}
\vspace{-8pt}
\caption{\textbf{Qualitative comparison with text-to-3D and text-to-avatar methods} (\S\ref{sec:sup_qua}). Text-to-3D methods for general objects are LatentNeRF \cite{metzer2023latent} and Fantasia3D \cite{chen2023fantasia3d}, and methods for avatars are AvatarCLIP \cite{hong2022avatarclip}, AvatarFusion \cite{huang2023avatarfusion} and AvatarBooth \cite{zeng2023avatarbooth}.}
\vspace{-12pt}
\label{fig:sup_qua}
\end{figure*}

We add comparison results with general text-to-3D generation methods, including LatentNeRF \cite{metzer2023latent}, and Fantasia3D \cite{chen2023fantasia3d}. Additional comparison with other text-to-avatar generation methods including Avatar-CLIP \cite{hong2022avatarclip}, AvatarBooth \cite{zeng2023avatarbooth} and AvatarFusion \cite{huang2023avatarfusion} are also provided. Comparison results of two avatars, ``Barack Obama" and ``Hillary Clinton" are shown in \cref{fig:sup_qua}.

General text-to-3D generation methods are likely to lose decent body shapes due to the lack of human prior knowledge. Even initialized with the human body, Fantasia3D still generates poor results because the normal guidance for geometry is unstable. Methods designed for text-to-avatar generation are better than general methods, but they still fail to control body proportions and maintain local structures. Compared with them, our method is able to generate photorealistic results with high resolution and high quality geometry and appearance.
\label{sec:sup_qua}

\section{Different diffusion models}
\begin{figure*}[t]
\centering
\includegraphics[width=0.95\textwidth]{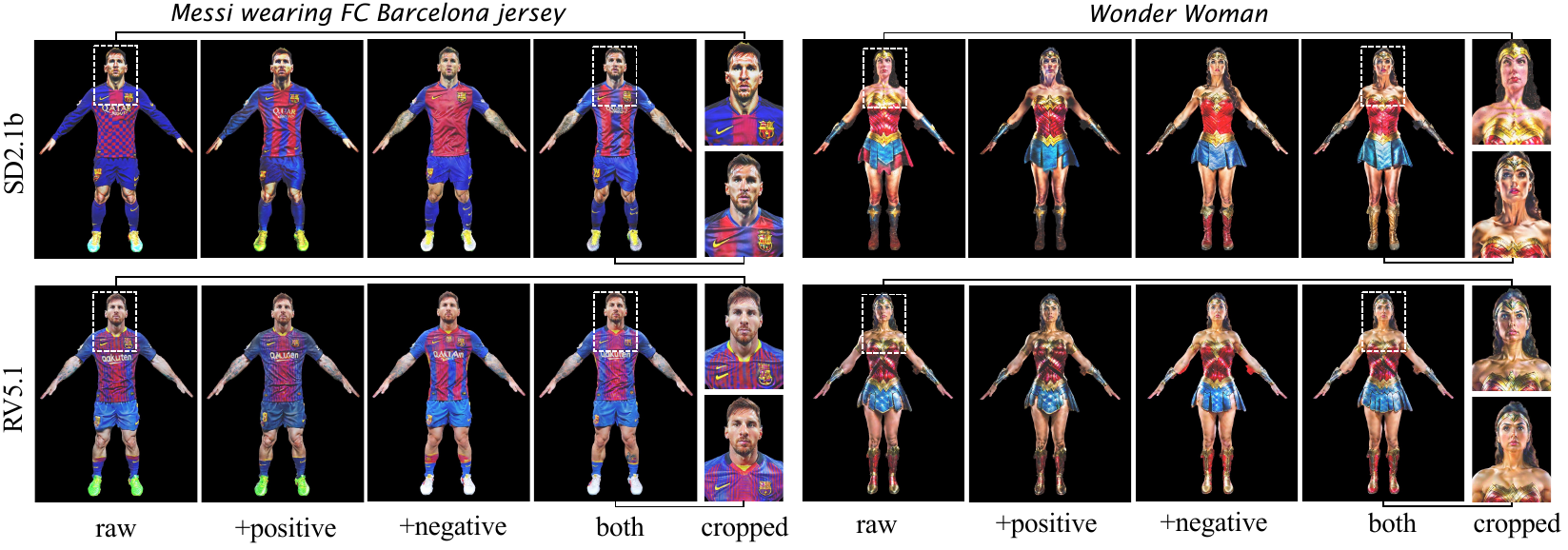}
\vspace{-8pt}
\caption{\textbf{Different diffusion models} for appearance generation (\S\ref{sec:sup_models}). Stable Diffusion 2.1 base (SD2.1b) and Realistic Vision 5.1 (RV5.1) are tested under different prompt strategies including raw prompt, adding postive prompt, adding negative prompt and adding both. Faces are croppped for better view.}
\vspace{-18pt}
\label{fig:sup_models}
\end{figure*}

For appearance generation, we use Realistic Vision 5.1 (RV5.1) \cite{RealisticVision} as the guidance diffusion model. Here we compare it with Stable Diffusion 2.1 base (SD2.1b) \cite{StableDiffusionv2.1-base}. First, we enhance both of the models with prompt engineering, and results are shown in \cref{fig:sup_models}. As for the raw results, RV5.1 is better than SD2.1b. If the positive prompt is used, both of the models yield richer details with higher contrast. If the negative prompt is used, the generation quality can be largely enhanced. Using both positive and negative prompt can produce results with rich details and high quality. However, RV5.1 has strong biases on female characters. The face of wonder woman in \cref{fig:sup_models} is not correct in the results of RV5.1, while SD2.1b can be used to make up for this shortcoming.

\begin{figure}[]
\centering
\includegraphics[width=0.8\columnwidth]{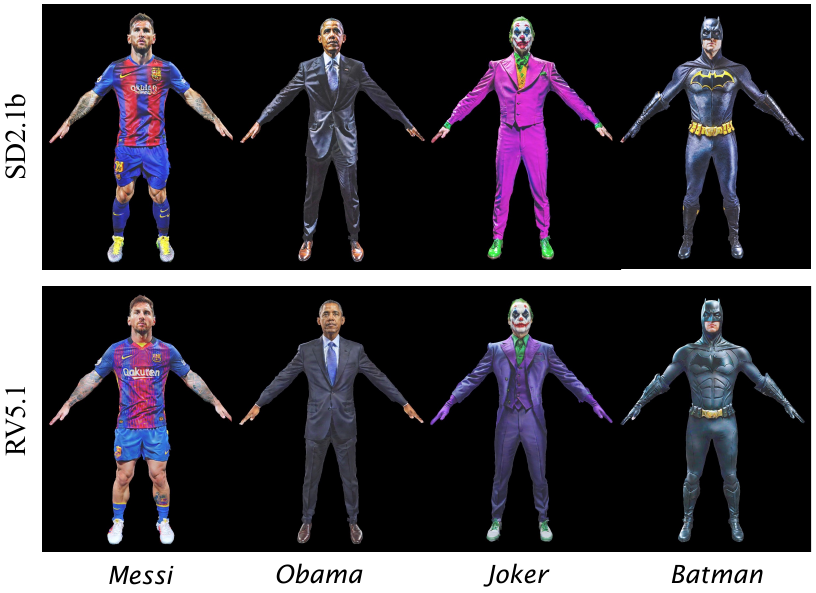}
\vspace{-8pt}
\caption{\textbf{Different color styles of diffusion models} for appearance generation (\S\ref{sec:sup_models}).}
\vspace{-12pt}
\label{fig:sup_styles}
\end{figure}

We also find that RV5.1 has better color styles compared with SD2.1b, as shown in \cref{fig:sup_styles}. RV5.1 shows more visually pleasing colors while SD2.1b may produce results with high contrast and saturation. Note the lightness constraint is not used in above results.
\label{sec:sup_models}

\section{Additional ablation study}
\label{sec:sup_ablation}
\begin{figure}[]
\centering
\includegraphics[width=0.95\columnwidth]{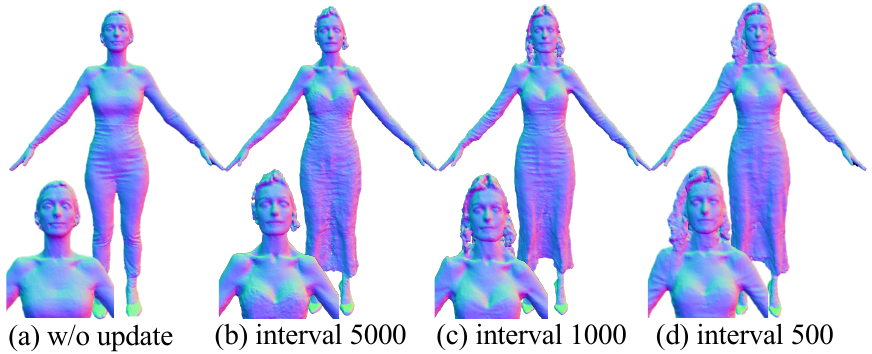}
\vspace{-8pt}
\caption{\textbf{Different settings for geometry evolving template} (\S\ref{sec:sup_ablation}). Detailed parts are cropped for better view.}
\vspace{-12pt}
\label{fig:sup_ablation_geo}
\end{figure}

For geometry generation, we present the results of different updating intervals of the evolving template in \cref{fig:sup_ablation_geo}. The prompt is ``Anne Hathaway,long wavy hair, wearing an ankle length dress". When the template is static human prior in (a), the global shape will be limited and the dress can not be generated. Once with an evolving template, the dress can be generated, while updating the template more frequently leads to more flexible shapes.

\begin{figure}[]
\centering
\includegraphics[width=0.95\columnwidth]{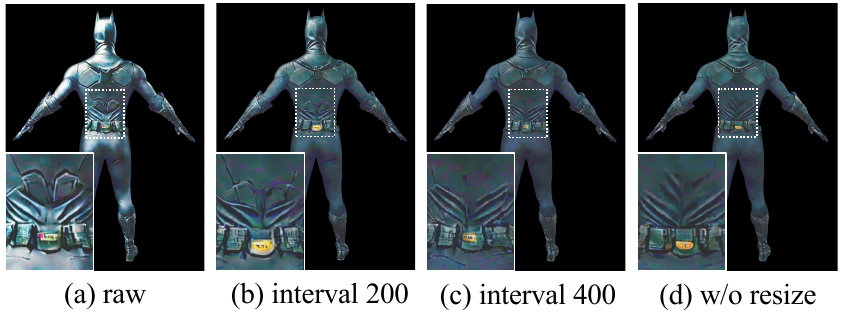}
\vspace{-8pt}
\caption{\textbf{Different settings for appearance evolving template} (\S\ref{sec:sup_ablation}). Detailed parts are cropped for better view.}
\vspace{-12pt}
\label{fig:sup_ablation_ap}
\end{figure}

For appearance generation, we present the results of different updating intervals and different image sizes for lightness constraint in \cref{fig:sup_ablation_ap}. A larger interval will help to suppress the lighting effect, while the details may also be affected. Similarly, applying lightness constraint directly on the raw size ($768\times768$) without down-sampling will also have stronger positive effect on suppressing lighting effect, but also more negative effect on details.

\section{Limitations and failure cases}
\label{sec:failure}
\begin{figure}[]
\centering
\includegraphics[width=0.95\columnwidth]{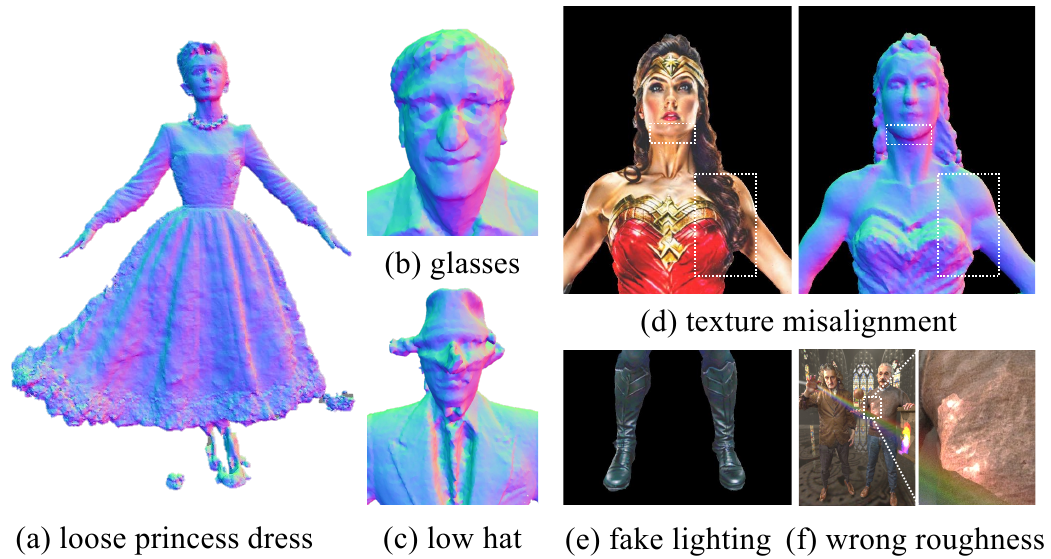}
\vspace{-8pt}
\caption{\textbf{Failure cases} (\S\ref{sec:failure}). (a,b,c) are for geometry and (d,e,f) are for appearance.}
\vspace{-12pt}
\label{fig:failure}
\end{figure}
We present some failure cases in \cref{fig:failure}. For geometry, (a) shows the case of generating a very loose princess dress. Since the dress is much deviated from the global human prior, which is beyond the capability of the evolving template, so our method can't generate it well. (b) and (c) show cases of local structures like glasses and hats. Generating these structures may be conflict with our local constraints from the human prior, so the results are not ideal.

(d) shows the misalignment problem between the geometry and appearance. Sometimes the appearance model fails to recognize correct geometry semantics and generates fake or wrong appearance. This problem may be alleviated by adding control in appearance guidance, like in \cite{liao2023tada}. (e) illustrates the fake lighting problem. Although our lightness constraint is able to suppress incorrect lighting effects, the problem is not perfectly solved. (f) shows the wrong specular caused by the wrong roughness. Our method is not able to generate accurate PBR parameters for materials. Introducing extra data of decoupled textures for training or finetuning may solve the issues better, like in \cite{xu2023matlaber}.

\end{document}